\documentclass[conference]{IEEEtran}
\usepackage{geometry}
\usepackage{cite}
\usepackage{amsmath}
\usepackage[numbers]{natbib}
\usepackage{amssymb}
\usepackage[parfill]{parskip}
\title{Detection, Recognition, and Tracking: A Survey}
\author{Shiyao Chen, Dale Chen-Song}
\date{\today}

\begin{document}
\maketitle
\section{Introduction}

For humans, object detection, recognition, and tracking are innate. These provide the ability for human to perceive their environment and objects within their environment. This ability however doesn't translate well in computers. In Computer Vision and Multimedia, it is becoming increasingly more important to detect, recognize and track objects in images and/or videos. Many of these applications, such as facial recognition, surveillance, animation, are used for tracking features and/or people. However, these tasks prove challenging for computers to do effectively, as there is a significant amount of data to parse through. Therefore, many techniques and algorithms are needed and therefore researched to try to achieve human like perception. 

In this literature review, we focus on some novel techniques on object detection and recognition, and how to apply tracking algorithms to the detected features to track the objects' movements.

\section{Detection}

Detection is a necessary initial step before recognition and tracking. It deals with detecting whether certain objects, such as humans, building, or cars, are in digital images and videos. Each object have their own special features that can be used to identify to classify such objects, such as shapes and lines. There are many problems that occurs when trying to detect objects, and several articles sought to find a way to detect objects or motion. 

\subsection{Hough Transform}
Hough transform is used to detect lines, circles or other parametric curves, patented by Paul V.C. Hough to recognize complex lines in photographs in 1962. It creates a separate parameter space and finds grouping in this parameter space to find lines and curves. Any edge detection algorithm can be used to detect the edges, however there may be some points missing due to error or noise. Each point from the original image is projected onto a set of points corresponding to possible lines or curves in the parameter space \cite{BALLARD1981111}, by using equations for line using \(y=mx+b\), or normalizing it to:
\begin{equation}
    r=x\cos\theta+y\sin\theta
\end{equation}
or for a circle, the parametric equation is \((x-a)^2 + (y-b)^2 = r^2\). 
\citet{fiala2002hough} applies such technique to detect horizontal line features in a non-single viewpoint (SVP) panoramic images, though the horizontal lines are not always straight, but curved. They map edge pixels to a new 2D parameter space for each mirror lobe to find the location of the existent horizontal line. Then, they use the panoramic Hough transform for non-SVP catadioptric systems. \citet{bernogger1998eye} also uses Hough transform to search for the irises in the eyes. The irises are the most significant feature of the eyes and is a simple shape, a circle. Even though there are many edgels, the Hough space will show a significant maximum at the exact position of the center of the iris. This detection of the iris and therefore the eyes, will be discussed further in Section IV as they then tracked the eyes. 

\subsection{Facial Feature Detection}
\label{sec:facial_detection}
As humans, we are programmed to see human faces quickly, even if it is in inanimate objects, which is called face pareidolia. However, face detection proves to be a task for computers. Face detection can recognize and locate facial features, get the contours of facial features and recognize facial expressions. There are many applications for face detection such as facial motion capture, facial recognition, lip reading and many more. But importantly, it is necessary for facial feature tracking, which will be further discussed in Section IV, and therefore a required step that needs to be implemented. 

\citet{yin2001generating} applied facial feature detection to generate realistic facial expressions with wrinkles. This would need to detect the necessary facial features for active wrinkle areas and mouth-eye areas to then synthesize facial expressions with wrinkles. Twenty-seven feature points are defined on the human face; the eyebrows, eyes, noses, mouth and hair. \citet{yin2001generating} generated a feature detection algorithm by estimating shapes to equate for each feature. For example, the eyes use Hough transform as mentioned above to detect iris as they are circles, the chins are two parabolas, and so forth. Another example is the detection of the nose. The detection of nose shape can be important for facial expression synthesis \cite{lijun2001nose}, similar to the last article discussed.

\citet{lijun2001nose} uses individual deformable templates to detect and track the nostril shape and nose-side shape. A two-stage region growing method is used to find facial feature regions. The feature of a nose can be extracted from the feature regions using pre-defined deformable templates. An important nose feature is the shapes of nostrils, which is a twisted pair curve with a leaf-like shape. 

\subsection{Motion Detection}
\label{sec:motion_detection}
Motion detection detects the change of in position of an object relative to its environment. In image sequences and videos, motion detection is important to track objects.
However, an important part for motion detection is to negate the movement from the background, as the background will provide a lot of unnecessary noise that doesn't relate to the detected object motion in focus. \citet{elnagar1995motion} detected motion using background constraints. The motion is detected by using a moving camera that can rotate and translate on planar surface and then computed a mapping that corresponds to pixels in successive images. The basic idea is that any point in the image should satisfy the background constraint if it's static, whereas a point that lies on an independently moving object will most likely not satisfy the constraint. So, it compensates for the apparent motion of the background of the scene caused by camera motion. Therefore, detecting moving objects becomes easy and can be used for both rotation and translation.

\subsection{Analysis}
In this section, we looked at several articles that detects objects. It is important to determine the line or shape required to detect the shape, and therefore the object, correctly. Furthermore, limiting the areas of detection can help mitigate errors and noise by looking at the right place at the start, as the algorithm doesn't waste time looking in areas where the object won't be. Hough transform can help edge detection by filling in the line if any noise or error disjoints the line. Background compensation is necessary for motion detection as it removes background noise that does not pertain to the object in focus. This shows that noise and error can affect detection and that eliminating said error and noise can help greatly detect the correct shape. The more robust a technique against error and noise, the better the technique and algorithm, which is why Hough transform and background compensation are used.

\section{Recognition}
While similar to detection, recognition provides a more detailed definition as recognition can determine what type of thing the object detected is. Recognition is important for finding and identifying objects in an image or video sequences. Therefore, many techniques can be deployed to classify patterns. Hence, in this section, several pattern recognition techniques will be introduced. 

\subsection{Radon Transform}
\label{sec:radon}
J. Radon first introduced the Radon transform (RT) in 1917. The RT of a function $f(x,y)$ can be defined as
\begin{equation}
    \Re_f(r,a) = \int\int f(x,y)\delta(r-x\cos\alpha-y\sin\alpha) dxdy
\end{equation}
where r is the shortest distance of a line to the origin, and $\alpha$ is the angle formed by the distance vector to the line. The delta function equals to a non-zero value only if the point lies on the line. When applying a RT to a function for given angles, RT will produce the projection along the given angles.

In various researches, Radon transform has been used in pattern recognition. The recognition algorithm usually follows the procedures \cite{signh2005visual} \cite{singh2005pose}
\begin{enumerate}
    \item Image acquisition and pre-processing
    \item Medial skeletonization
    \item Radon transform computation
    \item Feature extraction and clustering
    \item recognition decision.
\end{enumerate}

\subsection{A Multisensor Technique}
\label{sec:multisensor}
Smart sensor technology is also widely applied to pattern recognition. A more novel multisensor technique was also introduced recently. This new method uses multiple sensors placed at different angles, which prevent fusing sensor data.

The multisensor technique improves the real-time gesture recognition accuracy. Since each of the sensor has its own analysis to the hand gesture and the optimal gesture estimation will be selected by the pretrained support vector machine model, which increases the recognition accuracy\cite{rossol2016a}. 

\subsection{Analysis}
\citet{singh2005pose} applied the recognition method introduced in section \ref{sec:radon} to ground air traffic control gestures and other human pose recognition, the experiments showed an around 90\% accuracy on recognition human poses/gestures. Therefore, the approach shows its ability to classify certain patterns and can be apply to more fields. However, the method has its drawback that it cannot recognition noisy images with gestures. It is because the noise sources cause a larger Radon transformation leading to misclassification.

The new multisensor method in \ref{sec:multisensor} greatly increases the estimation accuracy compared to the accuracy of the conventional single sensor method. Nevertheless, this method can only classify 3 different hand poses. When the number of hand gestures increase, there is no guarantee that the algorithm could produce the estimation as good as the results proposed in the paper.

\section{Tracking}

Finally, tracking algorithms can use object detection and/or recognition to locate said object in image sequences and/or video. A simplified explanation of tracking is that after the object is detected, it is assigned an ID, and is then monitored as it moves through the frames. Applications of tracking occurs in areas such as traffic monitoring, robotics, and medical imaging. We further observe several applications of tracking in this section. 

\subsection{Feature-based Tracking Algorithms}
Feature based tracking algorithms can be static or dynamic. A static algorithm extracts features first in one frame, and then intends to map the feature points to the following frames\cite{Chetverikov1998TrackingFP}. \citet{Lucas1981AnII} proposed a dynamic feature tracking algorithm that iteratively find good matchings among consecutive frames. Later, \citet{Shi1994GoodFT} further improved Lucas and Kanade's algorithm by matching features for rotated, re-scaled or sheared objects. The algorithm first extracts some feature points in each feature window (FW), and then it iteratively finds the matching points in the next frame's feature window such that the sum of squared differences $E$ is minimized \cite{Shi1994GoodFT}. And $E$ is calculated using the equation:
\begin{equation}
    \begin{split}
        \min E & = \sum_{x=0}^{X-1}\sum_{x=0}^{Y-1}[A(x+\Delta x, y+\Delta y, t+\tau) \\
         & = - B(x,y,t)]^2w(x,y)
    \end{split}
\end{equation}
where $B(x,y,t)$ is the first frame at time $t$, $A(x+\Delta x, y+\Delta y, t+\tau)$ is the succeeding fram e at $t+\tau$ and $w(x,y)$ is the standard KLT average weighting function and 
\begin{equation}
  w(x,y)=\begin{cases}
    1, & (x,y)\in FW.\\
    0, & \text{otherwise}.
  \end{cases}
\end{equation}
However, using this weighting function, the tracking performance is poor if the image sequences is in noisy environments. Therefore, \citet{SINGH20051995} present a strategy that improves tracking performance of KLT by using Laplacian of Gaussian (LoG) and Gaussian weighting functions. This new method assigns weights to each pixels depending on their location within the feature window \cite{SINGH20051995}. The suggested weighting functions IS defined as:
\begin{equation}
    w_{Gauss}(x,y) = \begin{cases}
        \frac{1}{\delta \sqrt{2\pi}}e^{-\frac{x^2+y^2}{2\sigma ^2}} & (x,y)\in FW\\
        0, & \text{otherwise}.
    \end{cases}
\end{equation}
\begin{equation}
    w_{LoG}(x,y) = \begin{cases}
        \frac{1}{\pi \delta^4}[1-\frac{x^2+y^2}{2\sigma ^2}]e^{-\frac{x^2+y^2}{2\sigma ^2}}& (x,y)\in FW\\
        0, & \text{otherwise}.
    \end{cases}
\end{equation}
where $\sigma$ is the variance.

\subsection{Facial Feature Tracking}
\label{sec:facial_tracking}
Facial feature tracking, and therefore face tracking, is an important task in numerous applications that involves identifying humans, such as surveillance. To complete facial feature tracking, facial feature detection needs to be done beforehand.
\citet{lijun2001nose} tracks the nose feature, as discussed previously in Section \ref{sec:facial_detection}. The features are then placed onto a 3D wireframe model so it can be matched onto the individual face to track the motion using energy minimization.

\citet{bernogger1998eye} uses accurate localization and tracking of facial features for MPEG-4 systems. Detecting the eye features, using Hough transform for the iris (as mentioned in Section IIA) and exploiting color information, occurs before eye tracking. The algorithm determines 2 coarse region for the eyes initially. Then, the boundary of eyes and eye lids are obtained with color information. Eye tracking is similar to detection in this article, but the region of interest is restricted to use the information of the previous frame, such as the position and the size of the extracted eye in the previous frame. Furthermore, for animation of eye movements, the eye detection and tracking algorithms extract the contours of the iris and eye lids in each frame. The iris center can be matched using color, to make sure they are the same iris. 

This eye matching algorithm is further used by \citet{lijun1999integrting}, as they track a face. Initially, they need to use background compensation, as mentioned in Section \ref{sec:motion_detection}, as they use an active camera that can tilt or pan. Then, a facial motion detection algorithm is used, where the eyes are using the algorithm by \citet{bernogger1998eye} (1998) and the mouth is using a similar algorithm. Next, they can extract the facial features and put the moving face onto a 3D wireframe model adaptation.

\subsection{Analysis}
Both of the LoG and Gaussian weighting functions show better performance in tracking in a noisy environment than using the original KLT weighting function. Moreover, the computational time only increase 10\% even if the algorithm assigns weight to each pixel. Therefore, with LoG and Gaussian weighting functions, the tracking performance of objects in the video with noise can be much better. 

The algorithm of facial feature tracking introduced in \ref{sec:facial_tracking} shows its robustness because of its resistance to noises. Moreover, the researchers also animate the facial model based on the features extracted from the face. With the facial animation and tracking techniques, we can apply these methods to many areas in real life. However, there are some problems to these methods. First problem is that the method proposed by \citet{bernogger1998eye} requires high resolution facial model to reconstruct the face. Therefore, it may be time/computational power consuming in real life usage. Second, the algorithm by \citet{lijun1999integrting} assumes that the active camera can only tilt or pan. It does not take rotation movement into account. Overall, the algorithms for object tracking in sequences has shown a significant performance, and each algorithm has its advantages and disadvantages. In the future, the researchers can further improve their algorithm and try to solve the limitations of the algorithms.

\section{Analysis \& Conclusion}
There are many complicated techniques needed to track an object, especially when detection and/or recognition are required. However, these techniques can be combined together to produce a tracking algorithm, such as using Hough transforms used to detect eyes for facial feature detection \cite{bernogger1998eye} \cite{yin2001generating} \cite{lijun1999integrting}, and background compensation \cite{yin2001generating} \cite{elnagar1995motion}. Importantly, this shows that object detection is a crucial component to get correct and perfect, or else tracking does not function properly. Furthermore, the more resistant an algorithm is to noise and error, the better the algorithm will perform. Therefore, the creation of an algorithm that is resistant to more noise and error would be more suitable for real life applications. 

To summarize, tracking algorithms in computer vision and multimedia can use both object detection and recognition. Once the object is detected and recognized, it can then be tracked. However, since these are complicated tasks, many algorithms and techniques are used to help determine such tasks.

In this survey, we have presented some techniques used for object detection, recognition and finally tracking. This area is vast and we have only covered a small portion of such solutions. As more and more research is done, more techniques and algorithms are found and used for detecting, recognizing and tracking objects. Furthermore, pre-existing techniques can be improved and built on and used in newer algorithms, which we can see even in this small size of literature.

\bibliographystyle{IEEEtranN}
\bibliography{main}

\end{document}